%% file: glacier-BayesianUNet.tex
\documentclass{article}
\usepackage{spconf}
\usepackage[utf8]{inputenc}
\usepackage{microtype}
\usepackage{cite}
\usepackage{amsmath,amssymb,amsfonts}
\usepackage{algorithmic}
\usepackage{graphicx}
\usepackage{textcomp}
\usepackage{xcolor}
\usepackage{makecell}
\usepackage{booktabs}
\usepackage{pbox}
\usepackage{multirow}
\usepackage{bm}
\usepackage{amsmath}
\usepackage{hyperref}
\hypersetup{
    colorlinks,
    linkcolor={red!75!black},
    citecolor={blue!75!black},
    urlcolor={blue!75!black}
}
\usepackage{cleveref}
\crefname{section}{Sec.}{Sections}
\crefname{figure}{Fig.}{Figure}
\crefname{table}{Tab.}{Table}
\crefname{equation}{Equ.}{Equation}
\usepackage{subcaption}
\newcommand{\bx}{\bm{x}}
\newcommand{\bX}{\bm{X}}
\newcommand{\by}{\bm{y}}
\newcommand{\bY}{\bm{Y}}
\newcommand{\bW}{\bm{W}}
\newcommand{\bhW}{\bm{\hat{W}}}

\title{Bayesian U-Net for Segmenting Glaciers in SAR Imagery}

\name{%
	\begin{tabular}{c}Andreas Hartmann\textsuperscript{1*}\thanks{\textsuperscript{*}Andreas Hartmann and Amirabbas Davari contributed equally.}, 
	Amirabbas Davari\textsuperscript{1*}, 
		Thorsten Seehaus\textsuperscript{2}, 
		Matthias Braun\textsuperscript{2}, 
		Andreas Maier\textsuperscript{1},\\
		Vincent Christlein\textsuperscript{1}\end{tabular}
}
\address{%
		\textsuperscript{1} Department of Computer Science, Friendrich-Alexander University Erlangen-Nürnberg, Germany\\ 
	\textsuperscript{2} Department of Geography \& Geosciences, Friedrich-Alexander University Erlangen-Nürnberg, Germany 
}
\begin{document}
\maketitle

\begin{abstract}
    Fluctuations of the glacier calving front have an important influence over the ice flow of whole glacier systems.
    It is therefore important to precisely monitor the position of the calving front.
    However, the manual delineation of SAR images is a difficult, laborious and subjective task. Convolutional neural networks have previously shown promising results in automating the glacier segmentation in SAR images, making them desirable for further exploration of their possibilities. In this work, we propose to compute uncertainty and  use it in an Uncertainty Optimization regime as a novel two-stage process.
    By using dropout as a random sampling layer in a U-Net architecture, we create a probabilistic Bayesian Neural Network. With several forward passes we create a sampling distribution, which can estimate the model uncertainty for each pixel in the segmentation mask.
    The additional uncertainty map information can serve as a guideline for the experts in the manual annotation of the data. Furthermore, feeding the uncertainty map to the network leads to 95.24\,\% Dice similarity, which is an overall improvement in the segmentation performance compared to the state-of-the-art deterministic U-Net-based glacier segmentation pipelines. 
\end{abstract}

\begin{keywords}
Glacier segmentation, Bayesian deep learning, Convolutional neural networks, Image Segmentation
\end{keywords}

\section{Introduction}
Within the last years Convolutional Neural Networks (CNNs) have shown their huge potential in image classification and segmentation.
An important measure of glacier state is the position of the calving front (the position where the glacier ends and icebergs calve off) of marine or lake terminating glaciers. It can change its position  and affect the ice flow of the whole glacier system. A reduction of the buttressing forces or detachment from pinning points (bedrock) at the glacier terminus due to calving front recession, can destabilize the ice dynamics and lead to further calving front retreat~\cite{furst2016safety}. Thus, monitoring the front positions is of high interest since it can provide indicatory information for changes in ice dynamics and future development of the glacier.
Due to the remote location and huge spatial extension of the ice covered regions, glaciologists use various remote sensing data sets to map the position and changes of the glacier calving fronts~\cite{baumhoer2018remote}. Multi-spectral optical imagery or synthetic aperture radar (SAR) satellite acquisitions are typically analyzed. Cloud cover, shadows and polar night limit the availability of optical data, whereas SAR data is only affected by limitations due to its acquisition geometry and topography of the area of interest, causing effects like shadowing and layover. 
The calving front position is typically obtained by visual inspection and manual mapping~\cite{baumhoer2018remote}. The sea or lake surface next to the calving front is often covered by ice-melange, a mixture of sea ice, icebergs and open water. Even for trained analysts, it is often difficult to separate between the ice-melange and glacier area, since both surfaces can have quite similar surface textures. This makes the calving front mapping a difficult, laborious and subjective task~\cite{paul2013accuracy}. Based on edge detection, enhancement and image classification, different (semi-)automatic calving front mapping methods were developed (see \cite{baumhoer2018remote} for a more detailed review). Those methods show good results if there is an open water surface next to the calving front. However, for ice-melange covered areas, the performance is limited. 
CNNs have already shown very good performances in various image segmentation tasks. Some studies on applying CNNs for the separation of ice-melange and glacier were carried out in recent years. Mohajerani et~al.~\cite{mohajerani2019detection} used a U-Net CNN on multi-spectral Landsat imagery to map calving fronts of four glaciers in Greenland. The first studies using CNN and SAR imagery were carried out by Baumhoer et~al.~\cite{baumhoer2019automated} and Zhang et~al.~\cite{zhang2019automatically}.  

Pure CNNs lack a measure of the confidence of their predictions. Bayesian learning can bridge this gap by providing a tool to infer the uncertainty of the model prediction. 
Gal and Ghahramani~\cite{gal2016dropout} suggested a system for employing a Bayesian Neural Network (BNN) in deep learning applications by using dropout layers to approximate a Deep Gaussian Process.
This was successfully adapted for multi-class segmentation by Kendall et~al.~\cite{kendall2015bayesian}, leading to segmentation performance improvements of $2$-$3\%$ on a modified SegNet. Hiasa et~al.~\cite{hiasa2019automated} introduced a Bayesian U-Net for Muscoloskeletal Modeling also leading to significant segmentation improvements. 

In this work, we introduce the Bayesian U-Net to the state-of-the-art glacier segmentation framework in SAR imagery, and develop a 2-Stage pipeline that uses the model uncertainty to further improve the segmentation accuracy. 

\section{Methodology}\label{sec:methodology}
\subsection{Bayesian Neural Network}
Our uncertainty approach follows closely the suggested Bayesian U-Net by Hiasa et~al.~\cite{hiasa2019automated}, who themselves followed the proposal by Gal and Ghahramani~\cite{gal2016dropout}.
The BNN suggestion by Gal and Ghahramani employs dropout sampling as a Monte Carlo Estimation framework, called ``MC dropout''. They could prove that the dropout approximates Bayesian inference over the network weights \cite{gal2016dropout}.
We can use the BNN for predictions by running several forward pass samplings and approximating a deterministic CNN by averaging over them.
Additionally, by calculating the variance of the sample distribution, we obtain a measure of model uncertainty. A higher value means that the output is more liable to random perturbations from the dropout sampling and thus the model cannot reliably segment the affected area of the input image.
The proof from Gal and Ghahramani~\cite{gal2016dropout} for the dropout sampling approximation of a BNN is summarized below:

Given a training data set of images $\bm{X} = \{\bx_1, \dots, \bx_n\}$ and the corresponding label set $\bm{Y} = \{\by_1, \dots, \by_n\}$, a neural network can be represented by $p(y=c|\bm{x}) = \mathrm{Softmax}(f(\bx,\bW))$, where $\bm{x}$ is an unseen image, $y$ the output label of a pixel, and $c$ the class label.
A probabilistic description of the neural network could be:
\begin{equation}\label{eq:bayes-eq}
p(y=c|\bm{x, X, Y}) = \int p(y=c|\bx,\bW) p(\bW|\bX,\bY)d\bm{W}
\end{equation}
with $p(\bW|\bX,\bY)$ as the posterior distribution.
Gal and Ghahramani \cite{gal2016dropout} defined the distribution of dropout as $q(\bm{\hat{W}})$ with $ \bm{\hat{W}} = \bm{W} \cdot \mathrm{diag}(\bm{z})$, where $\bm{z} \sim \mathrm{Bernoulli}(\theta)$ and $\theta$ being the dropout rate.
They could prove that this approximates the posteriors distribution in Eq.~\ref{eq:bayes-eq} and the equation can then be approximated by minimizing the Kullbach-Leibler divergence as:
\begin{equation}
    \begin{split}
        p(y=c|\bx, \bX, \bY) \approx \int p(y = c|\bx, \bhW)q(\bm{\hat{W}})d\bm{\hat{W}}\\\approx \frac{1}{T} \sum_{t=1}^{T} \mathrm{Softmax}(f(\bx, \bhW)),
    \end{split}
\end{equation}
where $T$ is the number of dropout samplings. The corresponding variance is given by:
\begin{multline}
    \mathrm{Var}(y=c|\bx, \bX,\bY) \\\approx \frac{1}{T} \sum_{t=1}^{T} \mathrm{Softmax}(f(\bx,\bhW))^{T}\mathrm{Softmax}(f(\bx, \bhW)) \\- p(y=c|\bx, \bX,\bY)^{T}p(y=c|\bx,\bX,\bY).
\end{multline}
which is equal to the sample variance of $T$ stochastic forward passes through the BNN.

\subsection{Processing Pipeline}\label{pipeline}
\begin{figure}[t]
	\includegraphics[width=\columnwidth]{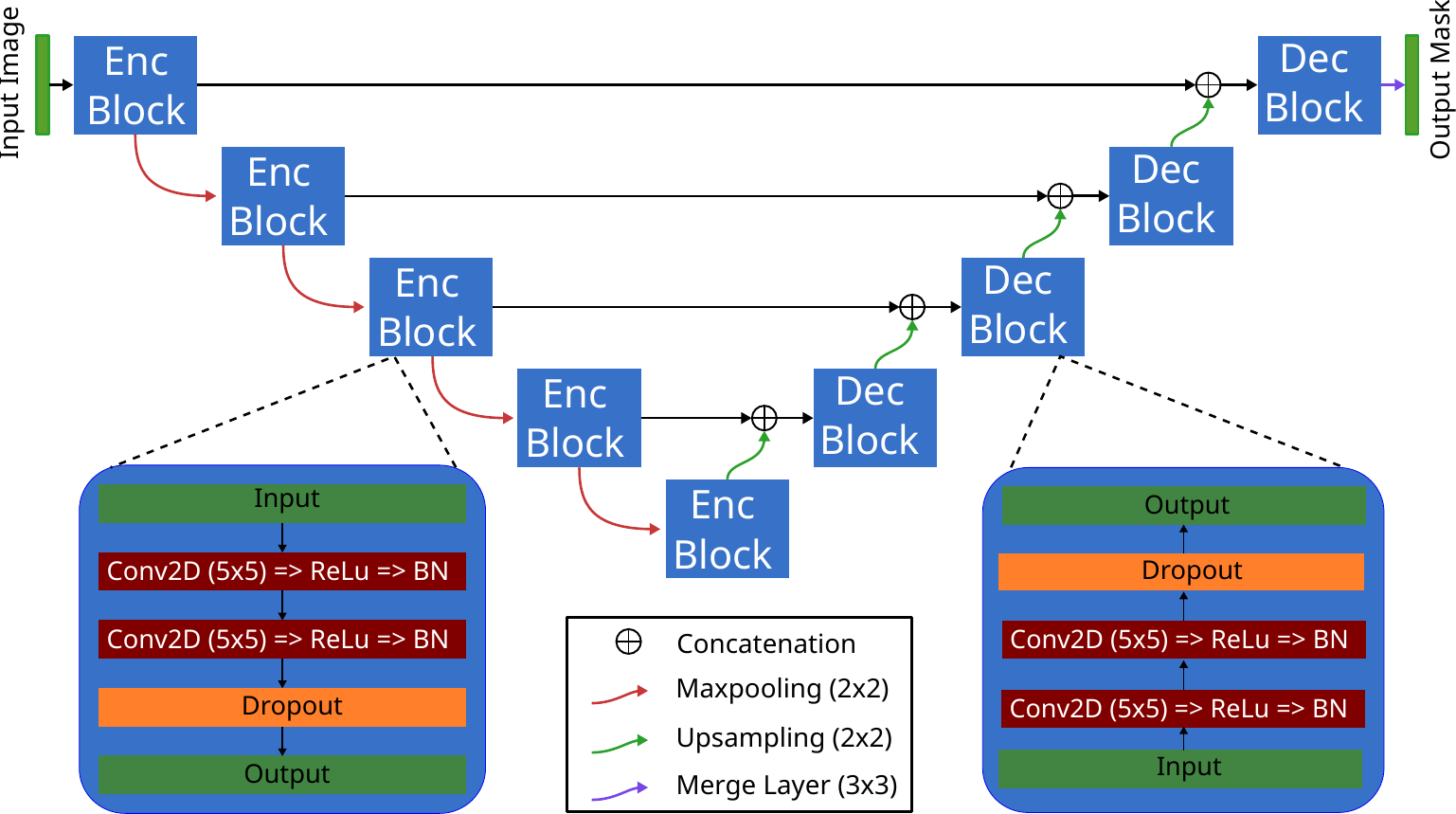}
	\caption{Model diagram of the Bayesian U-Net.}
	\label{fig:unet-model}
\end{figure}

The structure of our model is based on Zhang et~al.~\cite{zhang2019automatically}. We modified it by adding a dropout layer after every encoding and decoding block as suggested by Hiasa et~al.~\cite{hiasa2019automated}. The resulting Bayesian U-Net is displayed in \cref{fig:unet-model}. A fully Bayesian network should use a dropout layer for each convolution layer, however Kendall et~al.~\cite{kendall2015bayesian} found that this acted as a too strong regularizer, resulting in a lower training fit.

Our training process follows two steps:
In the first stage, we train a Bayesian U-Net with the images in the training subset and their corresponding segmentation ground truth maps. We then calculate an uncertainty map, by taking the variance of several forward passes of the trained network, using the training data as input. This map gives as an indication about the areas in which the model cannot reliably predict the segmentation output. We use this information in the next stage by training a second Bayesian U-Net.
With both the SAR-Image and the uncertainty map as a two-channel input, we train another Bayesian U-Net on the training data.
To get a segmentation output, several sampled forward passes are performed and the mean over the samples represents the segmentation prediction.


\section{Experimental Setup}\label{sec:experimental_setup}
~\\\noindent\textbf{Dataset.}
\noindent
In this study, we selected outlet glaciers on the Antarctic Peninsula to test and analyze the capabilities of CNNs for separating the ice-melange/water from the glacier on SAR intensity images.
At the Antarctic Peninsula, we selected the Sjögren-Inlet (SI) and Dinsmoore-Bombardier-Edgworth (DBE) glacier systems, which were major tributaries to the Prince-Gustav-Channel and Larsen-A ice shelves, respectively. 

Our dataset consists of the SAR imagery from the ERS-1/2, Envisat, RadarSAT-1, ALOS, TerraSAR-X (TSX) and TanDEM-X (TDX) missions, covering the period 1995-2014. In order to reduce speckle noise, the SAR imagery was multi-looked. Subsequently, the acquisitions were geo-coded, ortho-rectified using the ASTER digital elevation model from Cook et~al.~\cite{cook2012new}, and processed using GAMMA RS Software~\cite{gamma}. 
%
%
At DBE and SI glacier systems, we used the manually detected calving front locations from Seehaus et~al.~\cite{seehaus2015changes, seehaus2016dynamic}. 
%

~\\\noindent\textbf{Evaluation Protocol.}
\noindent
We randomly split the dataset into $144$ training, $50$ validation and $50$ test images. 
As established in \cref{pipeline}, we follow a 2-stage process. Both stages employ the same model architecture depicted in \cref{fig:unet-model}. 
The model takes $256\times 256$ grayscale image patches as input.
For each convolutional layer in the encoder and decoder blocks, we use a kernel size of $5\times 5$, starting with $32$ kernels at the uppermost block and ending with $512$ kernels for the last encoder block. Each convolutional layer is followed by a ReLU non-linearity and a Batch Normalization (BN) layer. We use a dropout layer after each block except the very last decoder block, each with a dropout rate of $0.5$. The extra channels are merged into a single channel output using a final $3\times3$ kernel convolutional layer.
%
The Bayesian U-Nets in both stages were trained using 
Adam optimizer~\cite{kingma2014adam} with an initial learning rate of $10^{-4}$. Furthermore, to avoid overfitting, we trained the models using early stopping with a patience of $30$ epochs and for a maximum of $250$ epochs.

The predicted uncertainty maps of the first stage were acquired by computing the variance over $20$ forward passes of the trained Bayesian U-Net. These uncertainty maps are binarized using a threshold of $0.125$. This value was calculated by anlyzing the $10$-bin histogram of the uncertainty values and selecting the border value of the last two bins.
Together with the SAR image patches, we used the uncertainty maps as a two-channel input to train the second stage Bayesian U-Net.
Afterwards, we computed the mean of $20$ forward passes to form the final predictions.


\section{Experimental Results}
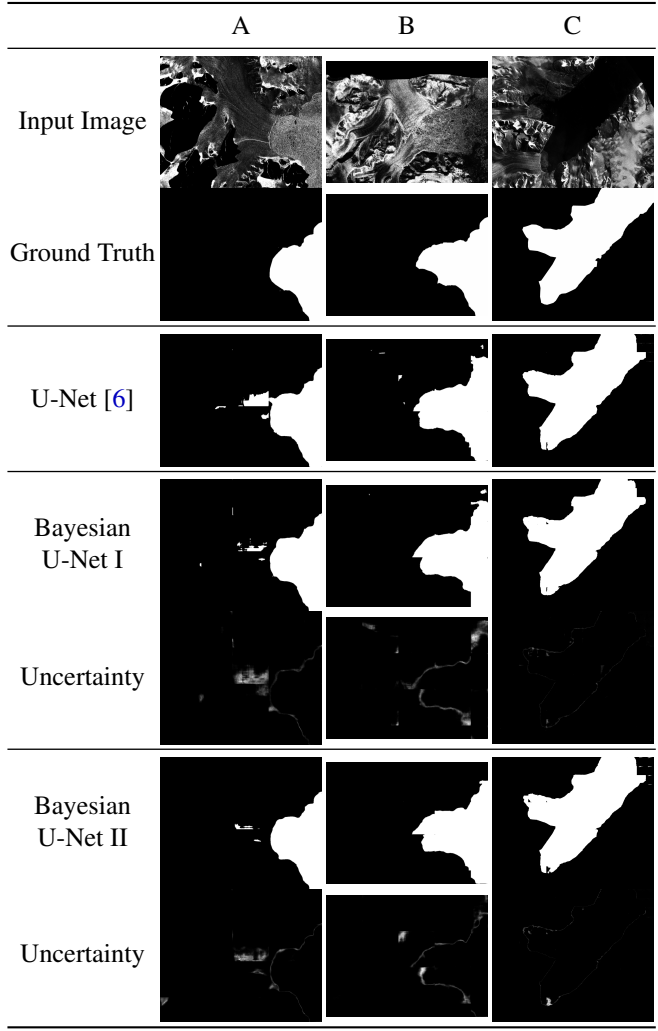
\begin{figure}[t]
\input{image_table}
\caption{Method comparison using three example images, A: ERS, B: ALOS, and C: TanDem-X. First two rows show the multi-looked SAR image and the manually segmented ground truth.
 Bayesian U-Net I denotes the prediction and uncertainty maps after the first stage while Bayesian U-Net II refers to the final prediction and the respective uncertainty map.}
\label{fig:image-table}
\end{figure}

We compare both our Bayesian U-Net and 2-Stage Uncertainty Optimization proposal with the state-of-the-art deterministic U-Net by Zhang et~al.~\cite{zhang2019automatically}. All networks are trained using the same parameters and datasets.
We evaluated the network for an image from each of the ERS, Envisat and TanDem-X platforms to show qualitative results depicted in \cref{fig:image-table}.
The network from Zhang et~al.~\cite{zhang2019automatically} already does a good segmentation of the SAR images. However it introduces some outliers in images A and B, and also fails to reproduce the fine details of the front line in the latter. 
Using the Bayesian U-Net (I), we can reproduce similar results with slightly different artifacts.
By looking at the now available uncertainty map, we can see that the uncertainty is high at the class transition, as expected. The other areas with high uncertainty correspond very accurately to the differences of the prediction and the ground truth. Using the second uncertainty optimization process (Bayesian U-Net II), we can use the extra information to reduce this error. 
The outliers are significantly reduced and some erroneous predictions in the homogeneous areas are corrected, leading to a more accurate segmentation.

\begin{table}[t]
	\caption{Evaluation Results for three example images and the complete test dataset. Images from A: ERS, B: Envisat, C: TanDem-X.}
	\label{tab:results}
	\resizebox{\columnwidth}{!}{
		\begin{tabular}{c c c c }
			\toprule
			Method & Image & Dice\% ($\pm$ SD) & IOU\% ($\pm$ SD) \\
			\midrule
			\multirow{4}{*}{U-Net~\cite{zhang2019automatically}} 
			& A & 93.28 & 87.41 \\
			& B & 92.58 & 86.19 \\
			& C & \textbf{96.71} & \textbf{93.63} \\
			& Complete Set & 94.86 ($\pm$ 6.21) & 90.22 ($\pm$ 3.20) \\
			\midrule
			\multirow{4}{*}{\makecell{Bayesian\\U-Net I (ours)}} 
			& A & 96.59 & 93.40 \\
			& B & 93.11 & 87.11 \\
			& C & 96.52 & 93.27 \\
			& Complete Set & 94.68 ($\pm$ 6.66) & 89.90 ($\pm$ 3.44) \\
			\midrule
			\multirow{4}{*}{\makecell{Bayesian\\U-Net II (ours)}} 
			& A & \textbf{97.55} & \textbf{95.22} \\
			& B & \textbf{95.01} & \textbf{90.50} \\
			& C & 96.57 & 93.34 \\
			& Complete Set & \textbf{95.24} ($\pm$ 6.57) & \textbf{90.91} ($\pm$ 3.40) \\
			\bottomrule
		\end{tabular}
	}
\end{table}

This can also be observed in the quantitative results shown in \cref{tab:results}. We report the Dice coefficient and Intersection Over Union (IOU), which are commonly used in evaluating the image segmentation performance.
For Image A and B the evaluation metrics follow the qualitative results. For image A, the Dice coefficient increases by 3.3\,\% after using the Bayesian U-Net. Further improvement by 1\,\% is obtained using the Optimized 2-Stage process. For image B the improvement is not as pronounced for the Bayesian U-Net, but the Uncertainty Optimization still holds its performance advantage over the deterministic U-Net. 
The Uncertainty Optimization produces the best results for both image A and B. The state-of-the-art U-Net and our Bayesian proposals perform comparably for image C. For the whole test set, all methods performed well in the segmentation task, with a mean Dice coefficient of circa 95\,\%. The first stage Bayesian U-Net and the deterministic state-of-the-art U-Net by Zhang et~al.~\cite{zhang2019automatically} perform comparably, while our proposed two-stage approach (Bayesian U-Net II) outperforms the other approaches and obtains the highest segmentation accuracy.



\section{Conclusion}\label{sec:conclusion}
We have presented a new deep learning-based glacier segmentation pipeline in SAR images, which relies on an uncertainty optimization process. Using a probabilistic Bayesian U-Net, we could calculate an uncertainty map, which approximates the model uncertainty for each pixel.
We could show that a high uncertainty reliably corresponds to misprediction, and could be used as additional information in a second Bayesian U-Net to improve the overall segmentation accuracy. With this 2-stage process, we could improve upon the deterministic state-of-the-art U-Net. This was especially successful in reducing false glacier classifications outside the main glacier body.
This also helps with automating the process of finding the glacier calving front since reducing those outliers reduces the need for extra filtering and likely leads to a cleaner output.
Finally, the uncertainty map that our pipeline provides could serve as a guideline and facilitate the manual annotation of the glaciers in SAR images by the experts for further training data generation. As our future work, we would like to investigate the effectiveness of our approach for this application.

\small
\bibliographystyle{IEEEbib}
\bibliography{references}

\end{document}

%% file: image_table.tex
\begingroup
\setlength{\tabcolsep}{1pt}
\begin{tabular}{cccc}
	\toprule
	& A& B&	C\\
	\midrule
Input Image
 &
  \pbox[c]{10em}{\includegraphics[width=0.12\textwidth]{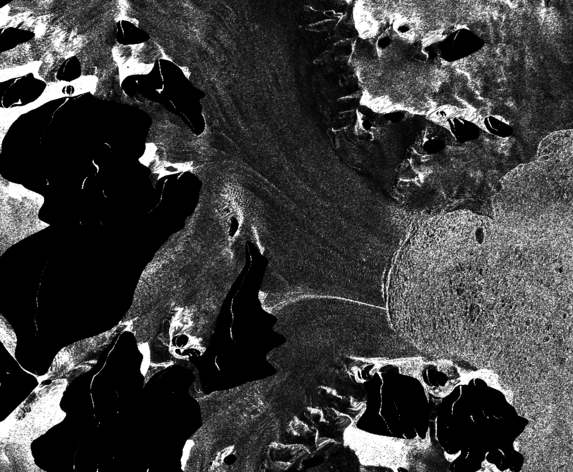}}
 &\pbox[c]{10em}{\includegraphics[width=0.12\textwidth]{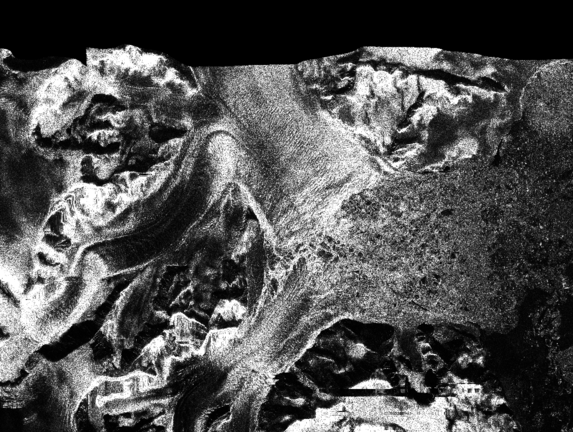}}
 &\pbox[c]{10em}{\includegraphics[width=0.12\textwidth]{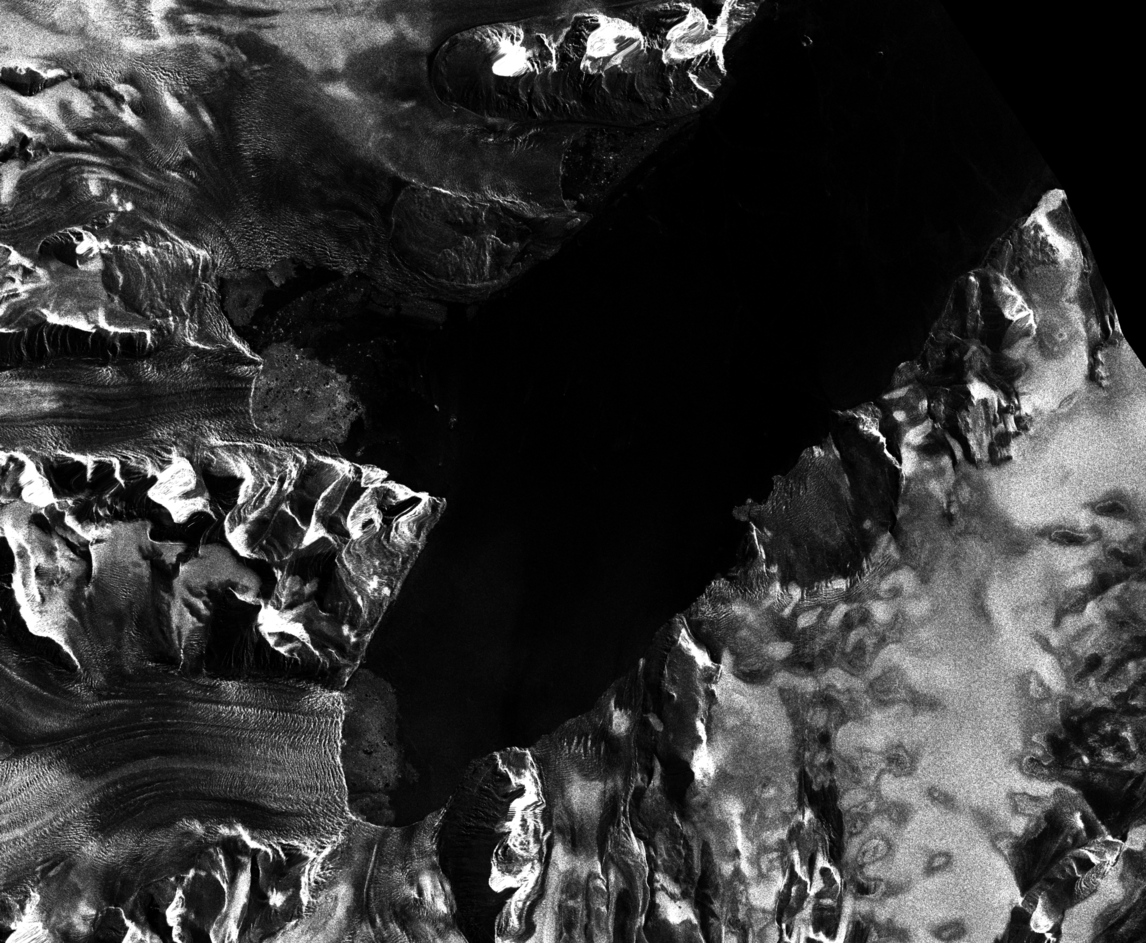}}
\\
Ground Truth
&
\pbox[c]{10em}{\includegraphics[width=0.12\textwidth]{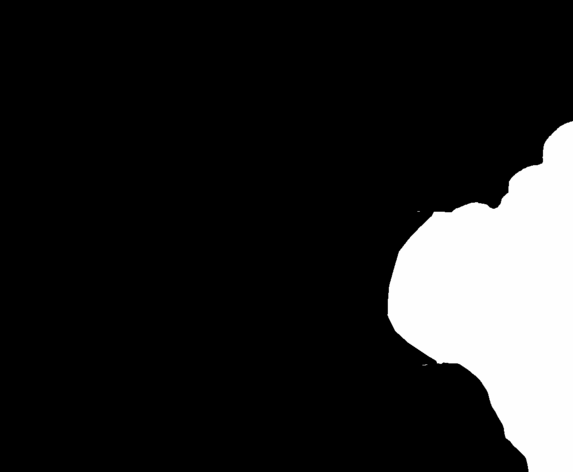}}
&
\pbox[c]{10em}{\includegraphics[width=0.12\textwidth]{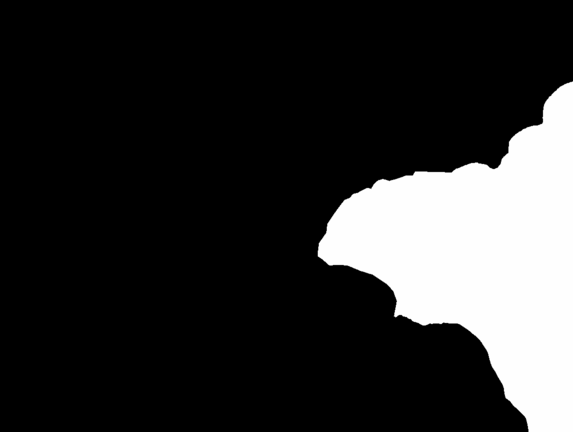}}
&
\pbox[c]{10em}{\includegraphics[width=0.12\textwidth]{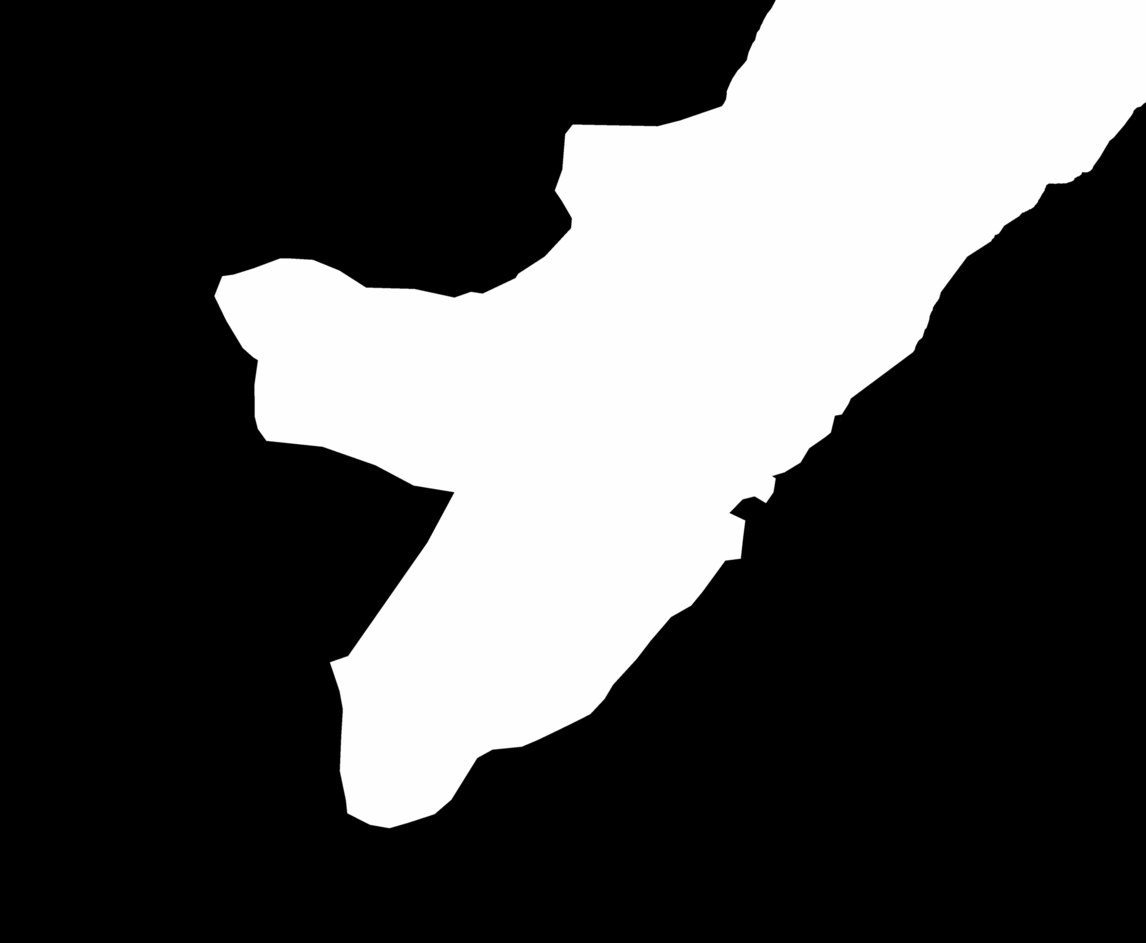}}
\\
\midrule
U-Net~\cite{zhang2019automatically}
&
\pbox[c]{10em}{\includegraphics[width=0.12\textwidth]{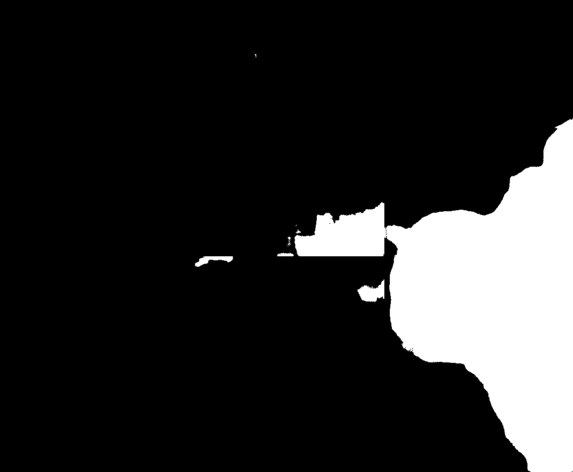}}
&
\pbox[c]{10em}{\includegraphics[width=0.12\textwidth]{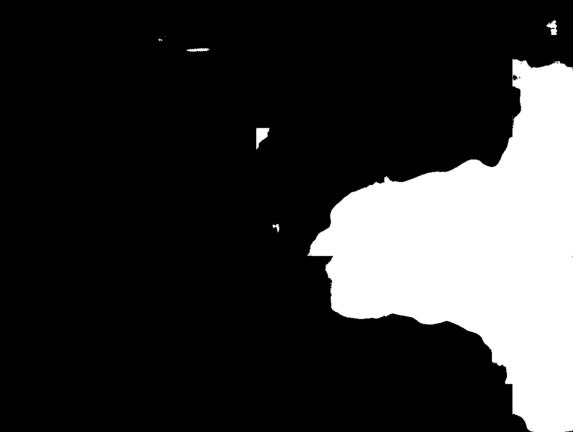}}
&
\pbox[c]{10em}{\includegraphics[width=0.12\textwidth]{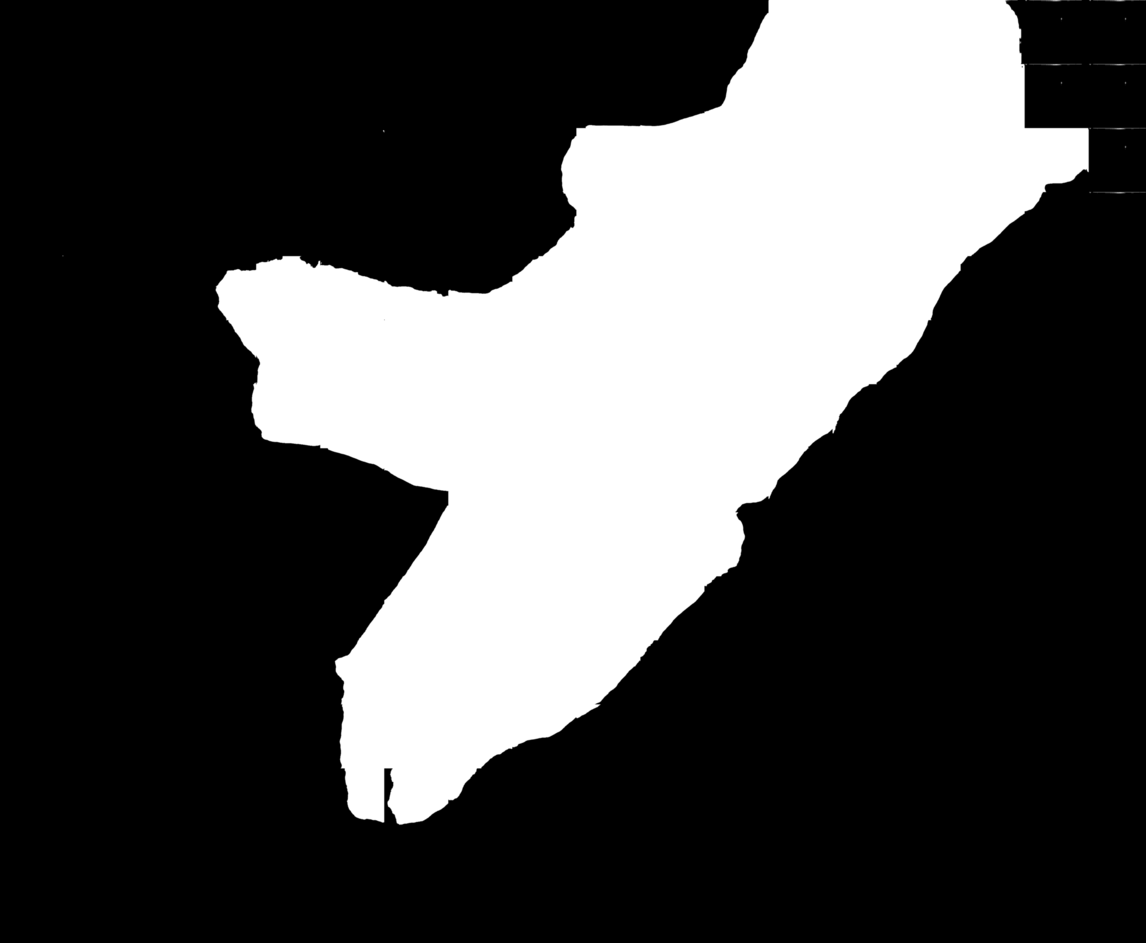}}
\\
\midrule
\begin{tabular}{c}Bayesian\\U-Net I\end{tabular}
&
\pbox[c]{10em}{\includegraphics[width=0.12\textwidth]{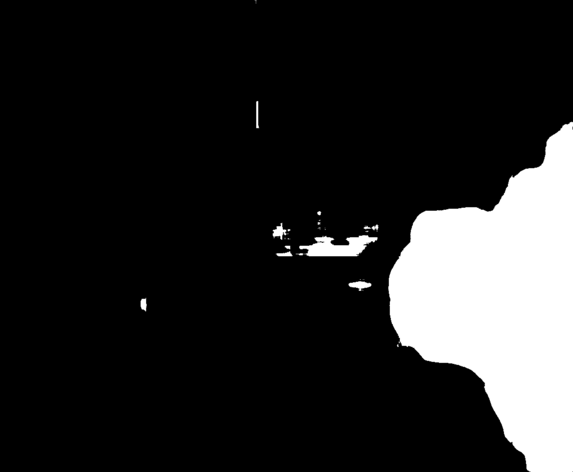}}
&
\pbox[c]{10em}{\includegraphics[width=0.12\textwidth]{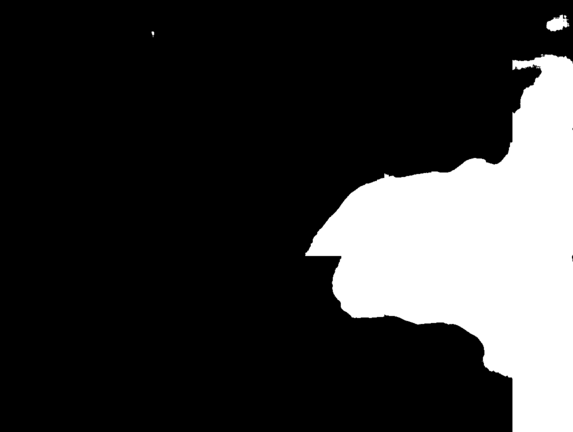}}
&
\pbox[c]{10em}{\includegraphics[width=0.12\textwidth]{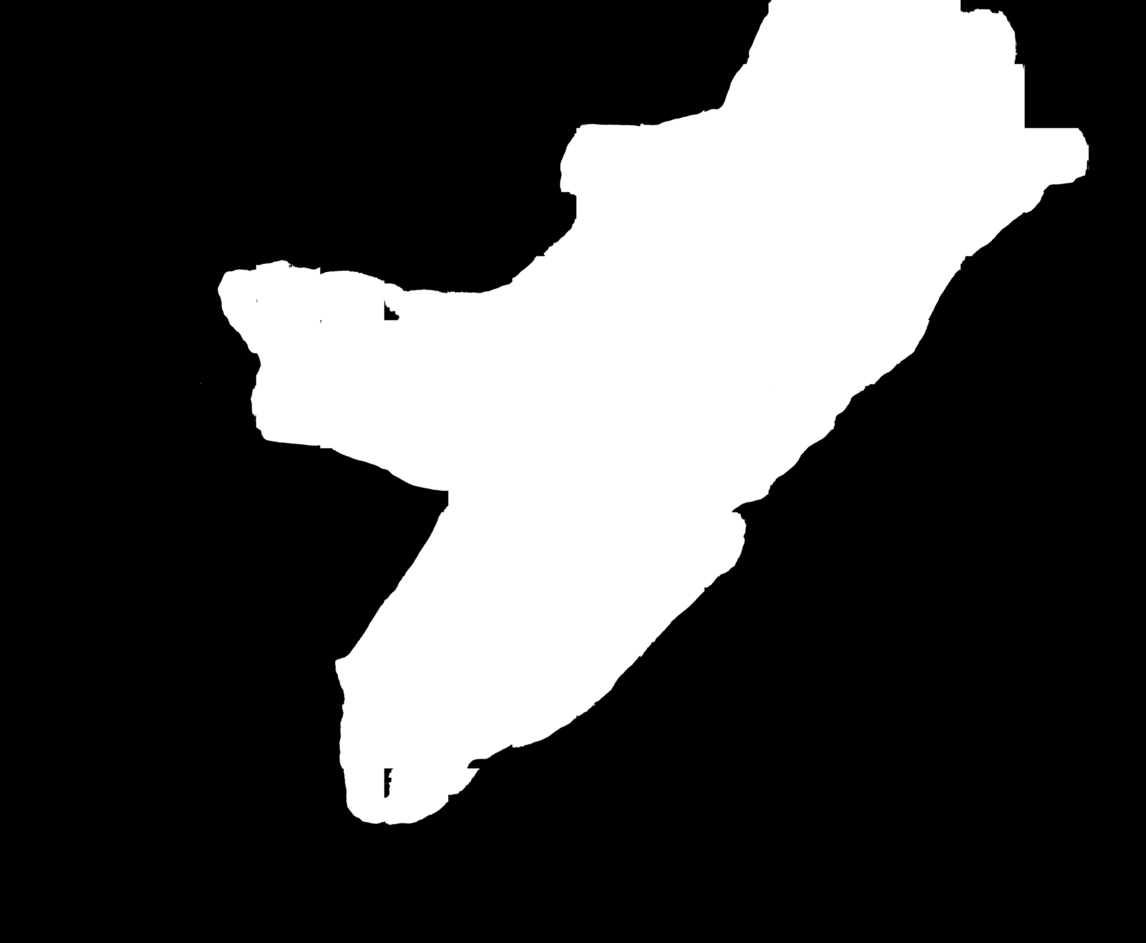}}
\\
 Uncertainty
&
\pbox[c]{10em}{\includegraphics[width=0.12\textwidth]{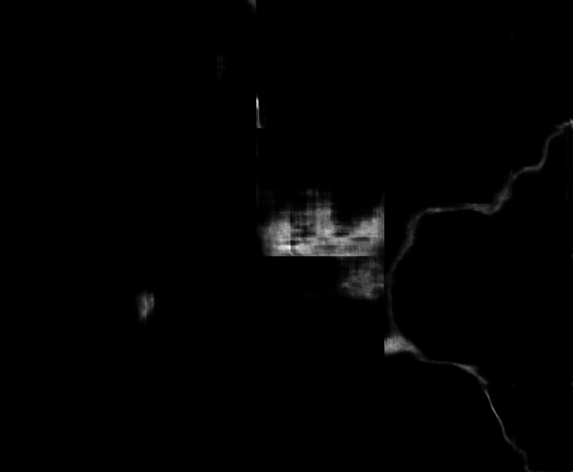}}
&
\pbox[c]{10em}{\includegraphics[width=0.12\textwidth]{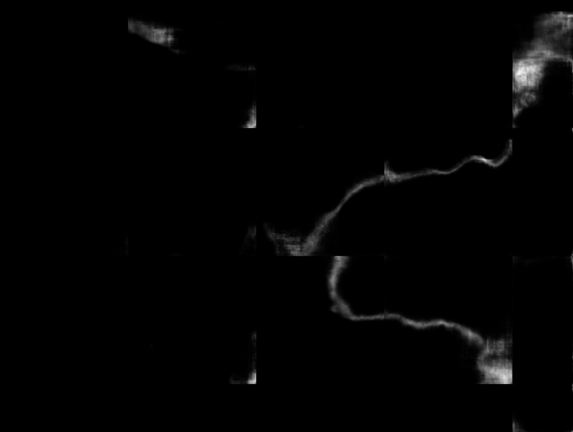}}
&
\pbox[c]{10em}{\includegraphics[width=0.12\textwidth]{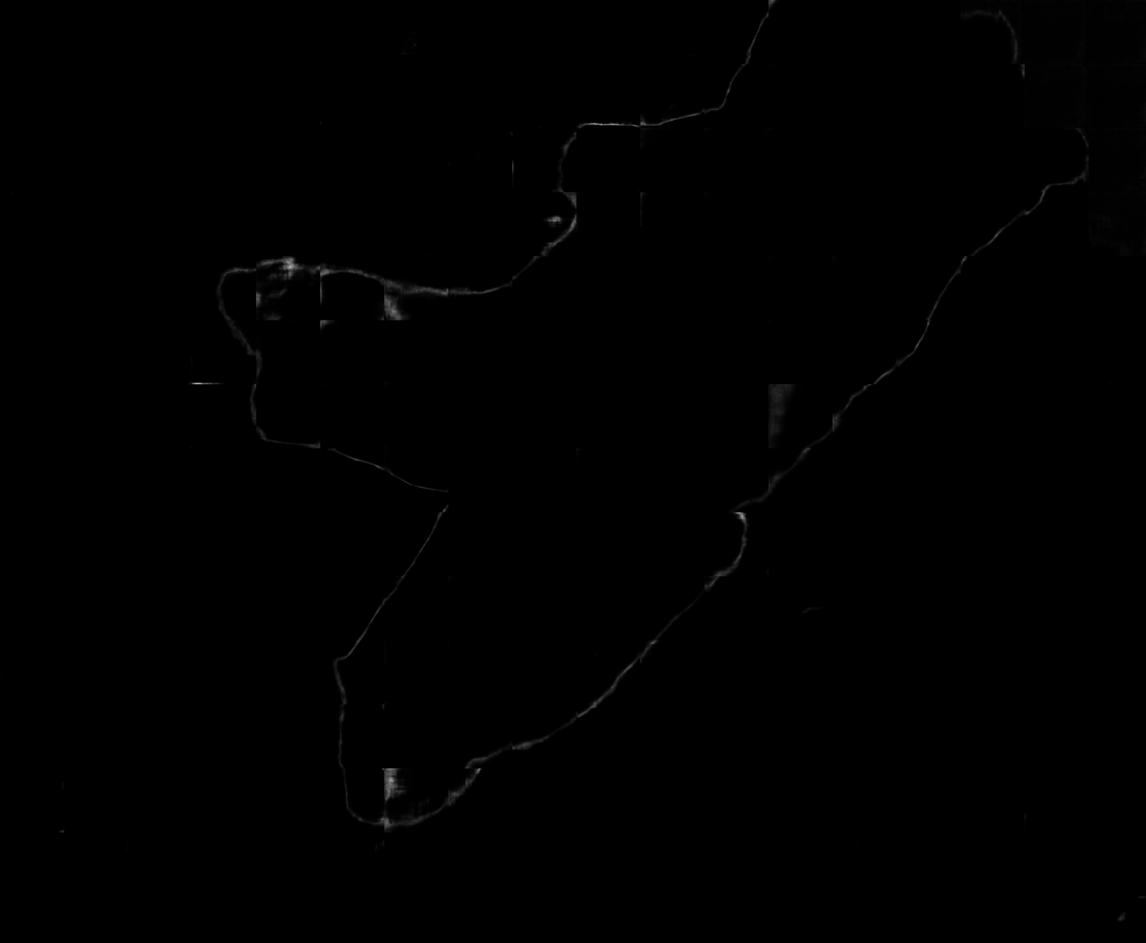}}
\\
\midrule

\begin{tabular}{c}Bayesian\\U-Net II\end{tabular}
&
\pbox[c]{10em}{\includegraphics[width=0.12\textwidth]{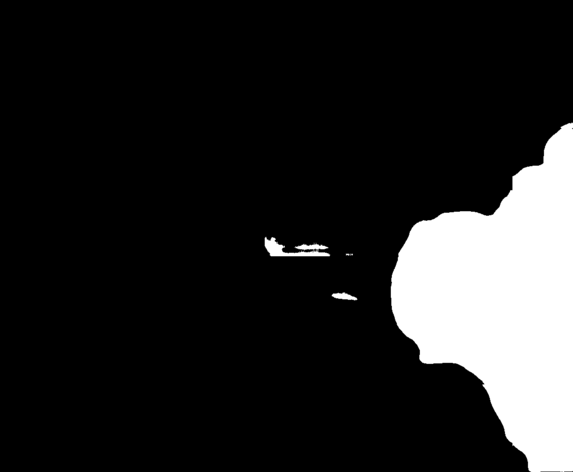}}
&
\pbox[c]{10em}{\includegraphics[width=0.12\textwidth]{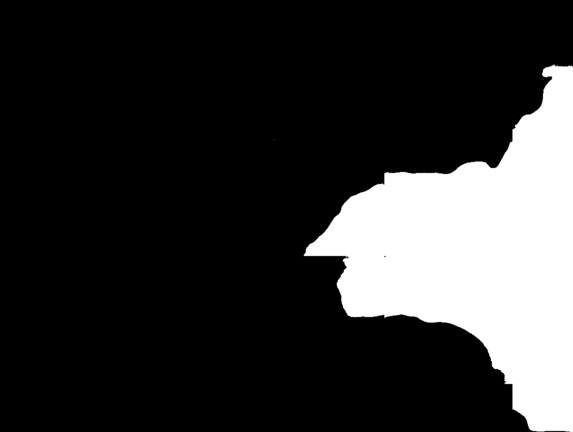}}
&
\pbox[c]{10em}{\includegraphics[width=0.12\textwidth]{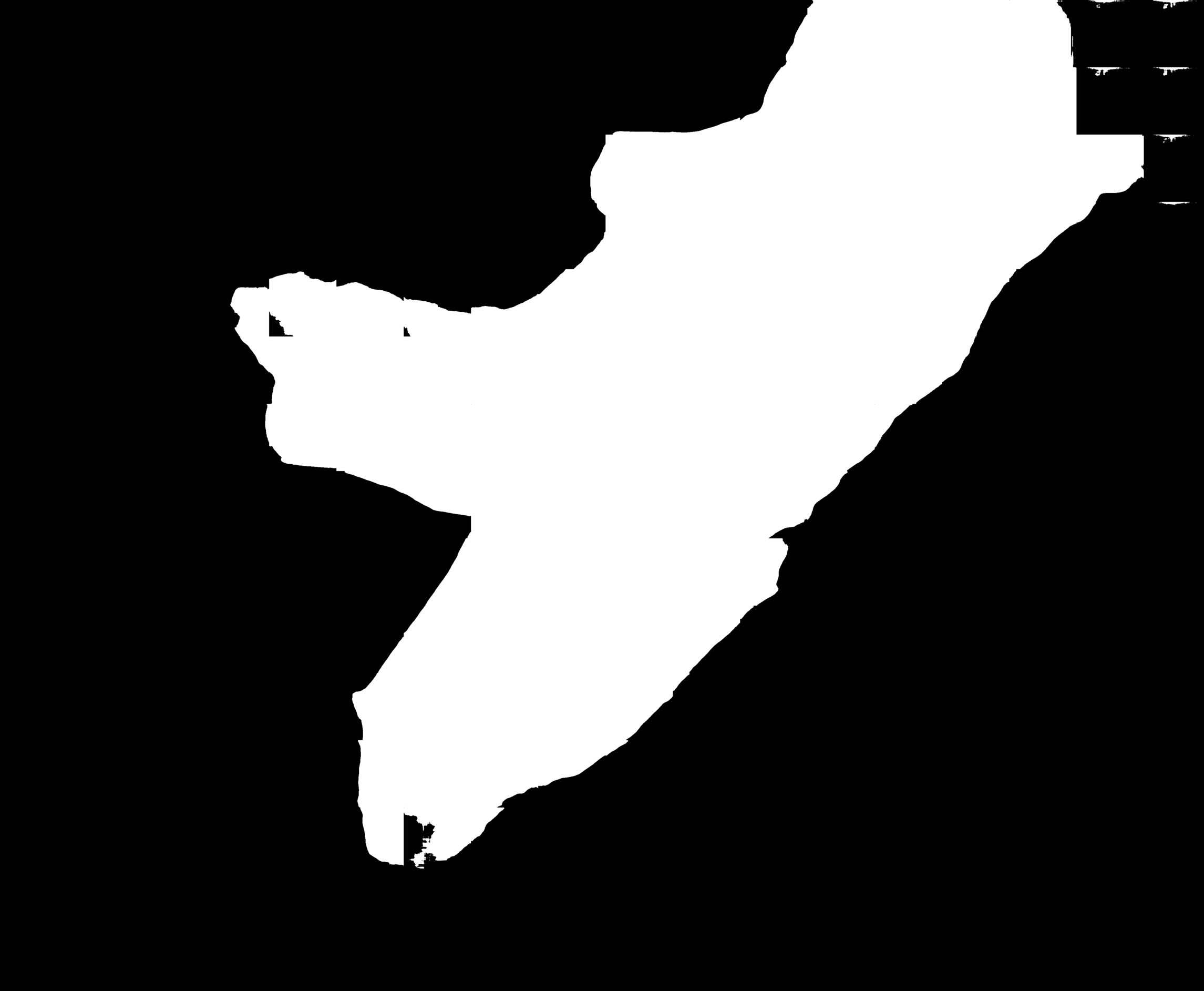}}
\\
 Uncertainty&
 \pbox[c]{10em}{\includegraphics[width=0.12\textwidth]{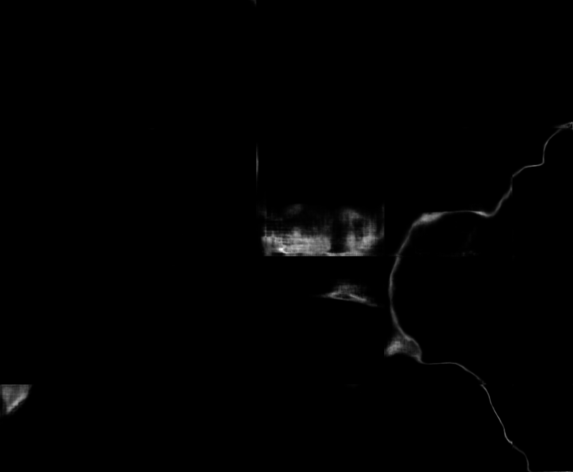}}
&
\pbox[c]{10em}{\includegraphics[width=0.12\textwidth]{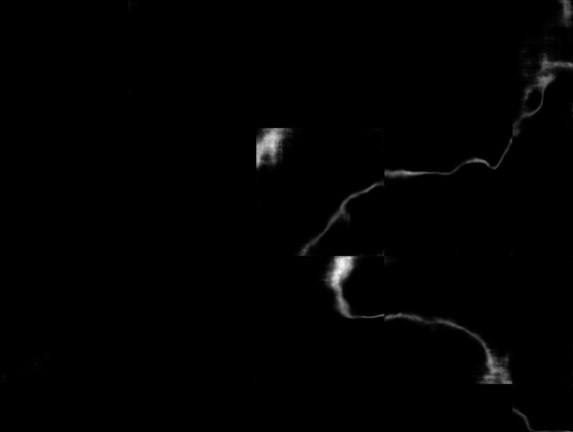}}
&
\pbox[c]{10em}{\includegraphics[width=0.12\textwidth]{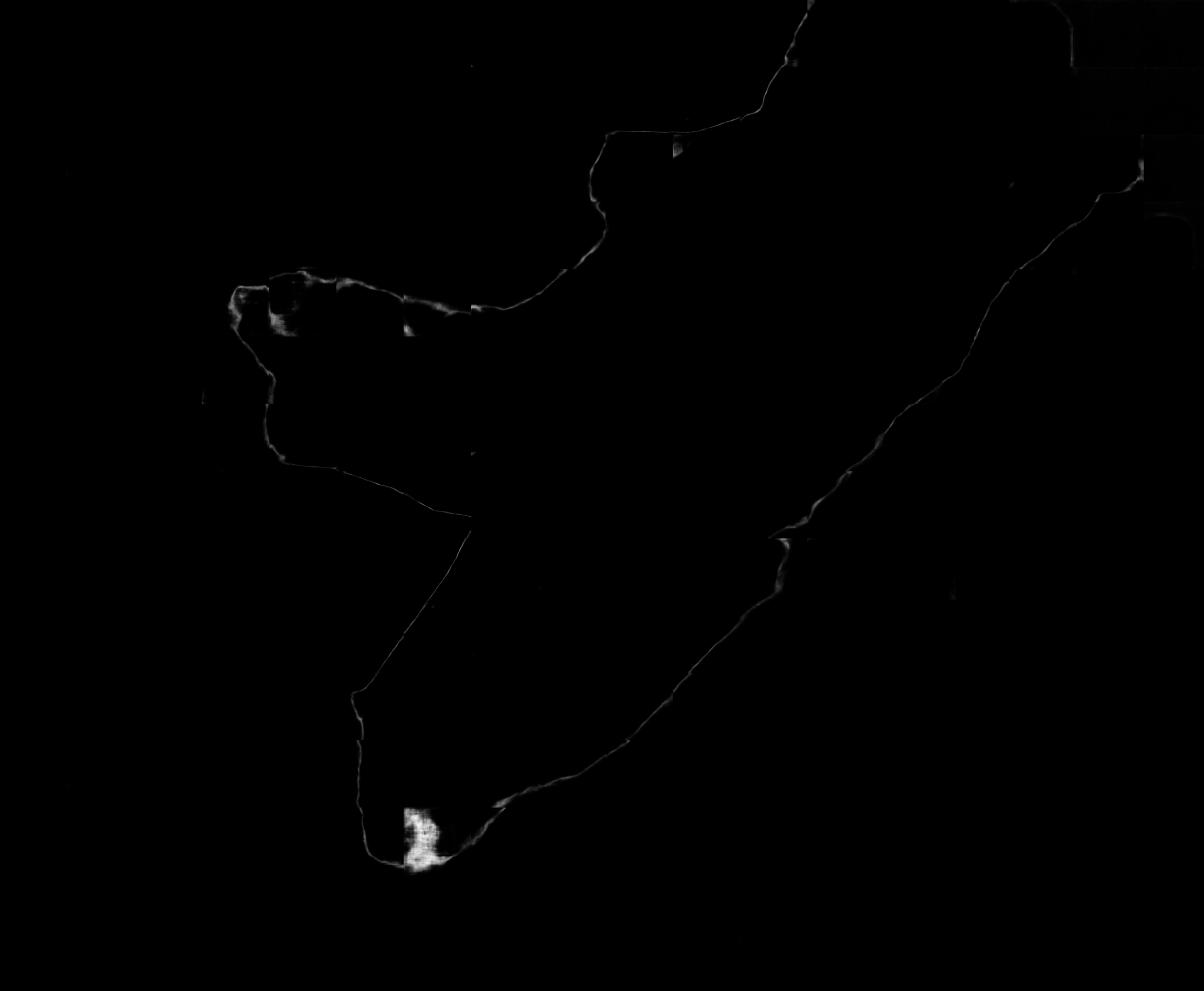}}
\\
\bottomrule
\end{tabular}
\endgroup